\pdfoutput=1
\documentclass[11pt]{article}

\usepackage[final]{coling}
\usepackage{amssymb}
\usepackage{booktabs} 
\usepackage{times}
\usepackage{latexsym}
\usepackage{hyperref}
\usepackage[T1]{fontenc}
\usepackage[utf8]{inputenc}

\usepackage{microtype}

\usepackage{inconsolata}
\usepackage{float}

\usepackage{graphicx}

\title{FActBench: A Benchmark for Fine-grained Automatic Evaluation of LLM-Generated Text in the Medical Domain}

\author{Anum Afzal \\
  Technical University of Munich \\
  \texttt{anum.afzal@tum.de} \\ 
  \And
  Juraj Vladika \\
Technical University of Munich \\
  \texttt{juraj.vladika@tum.de} 
  \AND
  Florian Matthes \\
  Technical University of Munich \\
  \texttt{matthes@tum.de} \\ \\}

\begin{document}
\maketitle
\begin{abstract}
Large Language Models tend to struggle when dealing with specialized domains. While all aspects of evaluation hold importance, factuality is the most critical one. Similarly, reliable fact-checking tools and data sources are essential for hallucination mitigation. We address these issues by providing a comprehensive Fact-checking Benchmark FActBench covering four generation tasks and six state-of-the-art Large Language Models (LLMs) for the Medical domain. We use two state-of-the-art Fact-checking techniques: Chain-of-Thought (CoT) Prompting and Natural Language Inference (NLI). Our experiments show that the fact-checking scores acquired through the Unanimous Voting of both techniques correlate best with Domain Expert Evaluation. 

\end{abstract}

\section{Introduction}

In the quickly evolving era of Natural Language Processing (NLP), Large Language Models (LLMs) are making their way into almost all use cases and domains. In most tasks, they have shown tremendous generative capabilities and a good understanding of text. However, they still tend to hallucinate in critical domains like the Medical domain. 
%When working on medical applications, practitioners are often faced with the decision of model choice. 
Contemporary LLMs are typically evaluated against general benchmarks and their assessment of the Medical domain is usually lacking. While it is essential to mitigate hallucinations, as a first step some reliable automatic fact-checking indicators are needed \citep{clusmann2023future}. The field of automatic fact-checking in LLMs is rapidly developing making it essential to find trustworthy techniques and data sources.

The state-of-the-art techniques for Automatic Fact Checking include Natural Language Inference (NLI) \citep{mor-lan-levi-2024-exploring,akhtar2024ev2revaluatingevidenceretrieval} using DeBERTa \citep{he-2020-deberta}, or through Chain-of-thought (CoT) \citep{Wei-CoT-2022} by using an LLM as a judge \citep{zheng2023judging}. Given the importance of Factual correctness in a critical domain such as medicine, it is helpful to rely on more than one technique for Fact-checking. Therefore, we explore the idea of Unanimous Voting such that an atomic fact is only considered to be factually correct if it is supported by both techniques.  

Hallucinations can generally be divided into input-conflicting, context-conflicting, and fact-conflicting \citep{zhang2023sirens}. The focus of our work lies in fact-conflicting, which is hallucination, where facts in output contradict the world knowledge. Additionally, our work builds on top of FActScore, a CoT-based approach for fact-checking. We adapt it to support user-provided grounding documents, making it suitable for tasks like RAG and Summarization. We present an automatic Fact-Checking Benchmark \textbf{FActBench}\footnote{Code for FActBench can be found at \href{https://github.com/jvladika/FactSumm/}{github.com/jvladika/FactSumm/}} with the following contributions:
\begin{itemize} 
    \item We fact-check six contemporary LLMs using Atomic Facts \citep{min-etal-2023-FActScore} on four generations tasks: Text Summarization, Lay Summarization, Retrieval Augmented Generation (RAG), and Open-ended Generation.
    \item We compare Intrinsic (Grounding Document) and Extrinsic (Wikipedia Dump) Fact-checking techniques in our experiments.
    \item We evaluate NLI, CoT as well as Unanimous Voting (UnVot) for the final prediction using domain expert evaluations as reference.
\end{itemize}

\noindent Details about all the datasets we use can be found in their original papers, including appropriate licenses and terms of use. 

\section{Related Work}
Hallucinations are a common problem in Natural Language Generation (NLG) tasks such as abstractive text summarization, generative question answering, or dialogue generation \citep{ji2023survey}. Detecting hallucinations is tied to the problem of measuring the factuality of model output \citep{augenstein2023factuality,zhao2024felm}. Hallucinations can be detected with approaches looking at the uncertainty in models' logits \citep{varshney2023stitch} or with approaches that fact-check model output over external knowledge sources \citep{chern2023factool}. 

Some recent works approached evaluation with question answering \citep{scialom2021questeval} or NLI \citep{utama2022falsesum}. Most recent methods leverage LLMs by querying them with prompts that directly ask for a score, like G-Eval \citep{liu2023g}, or evaluate the generated text with the veracity of its atomic facts, like FActScore \citep{min-etal-2023-FActScore}. \citet{fadeeva-etal-2024-fact} develop a method that does not require external knowledge for fact-checking as they leverage token-level uncertainty to identify the potentially factually incorrect generated section in the output. Similarly, \citet{sankararaman-etal-2024-provenance} introduces Provenance, a technique that uses NLI models to check if the RAG output is factually correct with reference to context. Lastly, \citet{chen2024factchdbenchmarkingfactconflictinghallucination} present FactCHD, a benchmarking for fact-conflicting hallucination detection for the General, Scientific, Health, and COVID-19 domains.

\begin{figure*}[h]
\centering
  \includegraphics[width=0.9\textwidth]{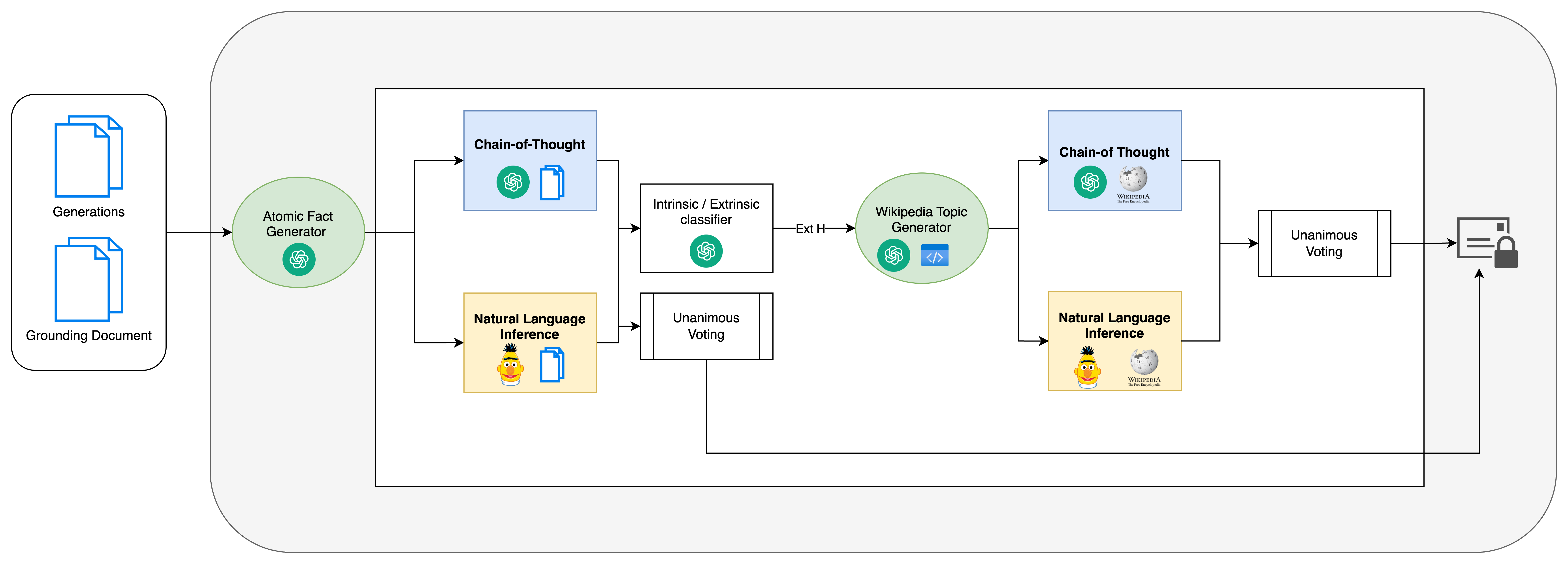}
  \caption{Block Diagram depicting how different fact-checking techniques interact with different data sources. Chain-of-Thought uses an LLM whereas Natural Language Inference uses a small LM as the backbone.}
  \label{fig:genfact}
\end{figure*}

\section{FActBench: Benchmark}

In our Benchmark, we use two SotA techniques, NLI and CoT, to evaluate 6 models on 4 different tasks. We follow the approach introduced by \citet{min-etal-2023-FActScore} to break all generations into a list of atomic facts which are then used for fact-checking. Since all our tasks with the exception of Open Generation, use a source document for grounding, we opt for a hybrid approach such that we first perform fact-checking using an intrinsic approach, followed by an extrinsic one.\footnote{In factuality evaluation, \textit{intrinsic} hallucinations are those that contradict the reference document, while \textit{extrinsic} hallucinations are those that contradict the external world knowledge.} The latter only performs evaluation on atomic facts that have been marked as hallucinations in the first step. We employ such an approach because it is possible for an atomic fact to be factually correct as per the world knowledge, even if it is not supported by the grounding document. We show this methodology in \autoref{fig:genfact} that illustrates how different fact-checking techniques and data sources interact with each other.

\subsection{Techniques}
\label{ssec:techniques}
\paragraph{Baseline: FActScore}
As a baseline, we first report on task performance using the established FActScore metric, following their external checks on Wikipedia with no grounding document. The reason we use it is its popularity in papers involving generative NLP tasks in the last couple of years \citep{dhuliawala-etal-2024-chain, chang2024survey, huang2025survey}. Later, our goal is to show that the combination of methods we use instead of raw FActScore lead to a more faithful evaluation framework and a better alignment with human scores.

\paragraph{Natural Language Inference (NLI):}
We utilize NLI as the first evaluation method. NLI aims to predict the logical relation between a premise and a hypothesis, including entailment, contradiction, and a neutral stance. We use the generated answer as the premise and the reference answer as the hypothesis. The intuition behind this approach is that a good answer should logically entail the reference. NLI has been applied for evaluating the quality of summaries and text generation \citep{mishra-etal-2021-looking, laban-etal-2022-summac, steen-etal-2023-little}.

Following this approach, we use DeBERTa-v3 \citep{he2023debertav}, shown to work well with NLI and reasoning tasks. We use the version \textit{Tasksource}, fine-tuned on a wide array of NLI \& classification datasets, which works well with long inputs \citep{sileo2023tasksource}.\footnote{\url{https://huggingface.co/tasksource/deberta-base-long-nli}} 
%This model predicts one of the three classes -- we
We take \textit{entailment} predictions as a sign of the atomic fact being supported by the original text and \textit{contradiction} as a sign of hallucination. We additionally check the contradicting atomic facts in an extrinsic way, by predicting their NLI class with the relevant Wikipedia context as the hypothesis. 
%If the prediction flips to \textit{entailment}, we update the decision, otherwise we keep it unchanged.

\paragraph{Chain-of Thought (CoT) Prompting:}
For evaluation using Chain-of-Thought Prompting, we adapted FActScore, an existing CoT-based fact-checking tool. This technique is suitable for open-ended generation and uses a Wikipedia dump as the knowledge source. FActScore supports extrinsic fact-checking by retrieving the most relevant passages from Wikipedia using user-defined topics. We adapt FActScore to support external documents as the basis for fact-checking. This "topic" should be the name of a real Wikipedia article, from which the relevant passages are retrieved. We also include a LLM-based topic generator so it is not required to manually define the topic when evaluating using passages from the Wikipedia dump. We use GPT-4o mini as the backbone of FActScore+, which serves as a compromise between cost and quality.

\paragraph{Unanimous Voting (UnVot):}
To produce a reliable fact-checking approach, we explore the idea of Unanimous Voting. This means we only consider an atomic fact to be correct if both NLI and CoT support it. This technique is especially useful for applications where high precision is needed.

\paragraph{Human Evaluation:}
We evaluate CoT, NLI, and UnVot techniques by correlating to domain expert judgment. We recruited 8 in-house employed individuals with a medical background to serve as annotators. A random subset of 80 generations (20 per task) was manually annotated such that each generation was evaluated by two annotators. They were instructed to follow the same hybrid, using both the original article and Wikipedia as a basis for fact-checking. Annotators were asked to assign a score between 1 and 100 to the generation estimating the factual correctness of the text.

\subsection{Tasks}
We include four tasks in our Benchmark, including Text Summarization, Lay Summarization, Retrieval Augmented Generation, and Open-ended Generation. The prompts used for all four tasks are shown in \autoref{app:prompts}. We summarize the datasets used for the tasks in \autoref{tab:dataset_size_analysis} and discuss them below. All the datasets can be found in their respective original papers, together with appropriate licenses.

\textbf{Text Summarization.} This task refers to the ability of an LLM to summarize a long scientific article into a summary. We used 1000 random samples from the PubMed Summarization dataset ~\citep{cohan-etal-2018-discourse}, which is derived from the original PubMed dump.

\textbf{Lay Summarization.} Contrary to normal text summarization, Lay Summarization refers to the model's ability to create a layman summary of biomedical articles. We use 1000 random samples from the PLOS dataset introduced by \citet{goldsack-etal-2022-laysumm}.% This dataset is similar to PubMed in nature but reference summaries are less technical explanations of otherwise complicated concept.% This tasks is especially useful in a fact-checking as the probability of LLM hallucinations can be higher if it trying to simplify difficult concepts.

\textbf{Retrieval Augmented Generation (RAG).} 
We use BioASQ-QA \citep{Krithara2023}, a biomedical question answering (QA) dataset
%benchmark dataset containing questions in English, along with gold reference answers and related sources. It has been 
designed to reflect the real information needs of biomedical experts. The questions are written by experts and evidence comes from PubMed.
%the large database of biomedical research papers. 
%The dataset is a part of the ongoing shared challenge with the same name and we use the 2023 version, Task 10b. 
%While the full dataset contains various types of questions,
We use the \textit{summary} subset -- 1130 questions paired with human-selected evidence snippets from PubMed and human-written "ideal answers" based on those snippets. We use the gold snippets as input to an LLM and prompt it to generate an answer to the given question, thus simulating a \textbf{RAG} pipeline.

\textbf{Open-ended Generation.}
%Unlike the first three tasks which are all based on input context, 
In this setting, no context is used and the model is prompted to generate an answer based on its knowledge. We again use the BioASQ dataset from the RAG task -- we take the 1130 questions and use them as input to an LLM by prompting it to answer the question.

\label{ssec:tasks}
\begin{table}[htpb]
\small
    \centering
    \begin{tabular}{lccc}
        \toprule
        \textbf{Task} & \textbf{Dataset} & \multicolumn{1}{c}{\#Source W} & \multicolumn{1}{c}{\#Gen W}\\
        \midrule
         Summ      & PubMed & 3,053.9 &256 \\
        Lay Summ  & PLOS & 6,696.8 & 256 \\
        RAG  & BioASQ-QA &351.9 & 116.5 \\
        Gen & BioASQ-QA& 351.9  & default \\
        \bottomrule
    \end{tabular}
\caption{Average word count of articles (\#W) and \# generation tokens (\#Gen W) during inference for tasks with respective datasets. Summ = Text Summarization, Lay Summ = Layman Summarization, RAG = Retrieval Augmented Generation, Gen = Open-ended Generation. }
\label{tab:dataset_size_analysis}
\end{table}

\begin{table*}[htpb]
\footnotesize
  \centering
  \resizebox{\textwidth}{!}{%
  \begin{tabular}{l|ccc|ccc|ccc|ccc}
    &  \multicolumn{3}{c}{\textbf{Summarization}} & \multicolumn{3}{c}{\textbf{Lay Summarization}} & \multicolumn{3}{c}{\textbf{RAG (QA)}}&\multicolumn{3}{c}{\textbf{Open-ended Gen}}\\
\cmidrule(lr){2-4}\cmidrule(lr){5-7}\cmidrule(lr){8-10}\cmidrule(lr){11-13}
    \textbf{Models} & \textbf{CoT} & \textbf{NLI} & \textbf{UnVot} & \textbf{CoT} & \textbf{NLI} & \textbf{UnVot}& \textbf{CoT} & \textbf{NLI} & \textbf{UnVot}& \textbf{CoT} & \textbf{NLI} & \textbf{UnVot}\\
\midrule
\multicolumn{13}{c}{\textit{FActBench (Grounding Document)}} \\
\midrule
    \texttt{GPT-4o mini}     & {95.8}& {77.4}  & {86.6}  & {95.4} & {94.8} &   {95.1}  &  {25.4} &  \textbf{77.7}  & {51.5} & {44.5} & \textbf{50.4} & {47.5} \\
    \texttt{Llama3.1~~8b}    & {95.3}& {87.8}  & {85.28}  & {95.4} & {93.5} &   {94.4}  &  {35.6} &  {76.7}  & {56.1} & {74.4} & {35.6} & {55.0}\\
    \texttt{Llama3.1~~70b}   & \textbf{96.52}& {84.59}  & {82.84}  & {96.1} & {94.1} &   {95.1}  &  {31.3} &  {76.3}  & {53.8} & {37.3} & {46.5} & {41.8}\\
    \texttt{Mistral~~7b}     & {95.8}& {82.55}  & {80.38}  & {96.3} & {97.32} &  {94.75}  &  {82.9} &  {73.1}  & {78.0} & {80.9} & {32.5} & {56.6}\\
    \texttt{Mixtral~~8 x 7b} & {95.2}& {87.86}  & \textbf{95.5}  & {96.5} & {97.0} &   {95.0}  &  \textbf{88.2} &  {75.0}  & \textbf{81.6} & \textbf{85.4} & {36.9} & \textbf{61.1}   \\
    \texttt{Gemma~~9b }      & {84.55}& {71.95}  & {68.77}  & {82.94} & {80.65} &   {75.48}  &  {35.8} &  {44.0}  & {43.7} & {54.1} & {30.5} & {28.0}  \\
      \midrule
\multicolumn{13}{c}{\textit{FActBench (Grounding Document + Wikipedia)}} \\
\midrule
    \texttt{GPT-4o mini}     & {96.8}& {82.6}  & {80.4}  & {96.2} & {96.6} &   {93.4}  &  {97.3} &  \textbf{78.2}  & \textbf{76.4} & \textbf{95.8} & \textbf{51.4} & \textbf{50.3} \\
    \texttt{Llama3.1~~8b}    & {96.4}& {88.85}  & {86.25}  & {96.5} & {94.2} &   {91.5}  &  {98.2} & {77.1} &  {76.1}  & {79.3} & {36.7} & {32.1}\\
    \texttt{Llama3.1~~70b}   & {97.27}& {85.71}  & {83.9}  & {97.0} & {94.8} &   {92.0}  &  {97.2} &  {76.8}  & {75.1} & {90.9} & {47.7} & {45.9}\\
    \texttt{Mistral~~7b}     & {96.51}& {83.59}  & {81.34}  & \textbf{97.83} & \textbf{96.7} &   \textbf{94.93}  &  \textbf{98.6} &  {73.5}  & {72.7} & {92.1} & {33.2} & {31.9}\\
    \texttt{Mixtral~~8 x 7b} & {96.9}&  {88.68}  & \textbf{86.24}  & {97.5} & {97.2} &   {95.1}  &  {97.7} &  {75.3}  & {74.0} & 93.0 & {37.8} & {36.5}   \\
    \texttt{Gemma~~9b }      & {93.03}& {74.46}  & {70.99}  & {91.11} & {81.68} &   {76.43}  &  {97.4} &  {45.0}  & {44.6} & {80.1} & {31.5} & {28.8}  \\
\midrule
\multicolumn{13}{c}{\textit{Baseline: FActScore (Wikipedia)}} \\
\midrule
\texttt{GPT-4o mini}     &\multicolumn{3}{c}{51.34}&\multicolumn{3}{c}{52.6}& \multicolumn{3}{c}{19.4}&\multicolumn{3}{c}{41.4}\\
\texttt{Llama3.1~~8b}    &\multicolumn{3}{c}{43.97}&\multicolumn{3}{c}{49.4}& \multicolumn{3}{c}{25.3}&\multicolumn{3}{c}{71.3}\\
\texttt{Llama3.1~~70b}   &\multicolumn{3}{c}{50.08}&\multicolumn{3}{c}{48.8}& \multicolumn{3}{c}{24.0}&\multicolumn{3}{c}{34.8}\\
\texttt{Mistral~~7b}     &\multicolumn{3}{c}{46.11}&\multicolumn{3}{c}{50.02}& \multicolumn{3}{c}{61.1}&\multicolumn{3}{c}{78.4}\\
\texttt{Mixtral~~8 x 7b} &\multicolumn{3}{c}{49.71}&\multicolumn{3}{c}{51.00}& \multicolumn{3}{c}{\textbf{64.5}}&\multicolumn{3}{c}{\textbf{81.6}}\\
    \texttt{Gemma~~9b }      &\multicolumn{3}{c}{\textbf{53.54}}&\multicolumn{3}{c}{\textbf{54.56}}& \multicolumn{3}{c}{44.0}&\multicolumn{3}{c}{52.0}\\
    
     \midrule
    
  \end{tabular}
  
  }
  \caption{Factchecking scores of six LLMs on four tasks using Chain-of-Thought (CoT) prompting, Natural Language Inference (NLI), and Unanimous voting (UnVot). We show scores by incorporating two different knowledge sources.}
  \label{tab:faithfullness}

\end{table*}

\subsection{Models}
\label{ssec:models}
We include six LLMs in our experiments including \texttt{Llama3.1 8b} \citep{dubey2024llama3herdmodels} \texttt{Llama3.1 70b}, \texttt{Mistral 7b} \citep{jiang2023mistral7b}, \texttt{Mixtral 8x7b} \citep{jiang2024mixtralexperts}, \texttt{Gemma2 9b} \citep{gemmateam2024gemma2improvingopen} and lastly, closed-source \texttt{GPT-4o mini}. We provide the checkpoints and technical details in \autoref{sec:appendix-training-details}.
%While domain-specific LLM generated text can be evaluated against various aspects \cite{afzal2024adaptevalevaluatinglargelanguage,liu-etal-2023-g}, we strictly focus on faithfulness in our evaluation. We prioritize faithfulness over other aspects since most contemporary LLMs do manage to generate plausible summaries however, are not always faithful \cite{agarwal2024faithfulnessvsplausibilityunreliability}. We use GenFAct to evaluate LLM-generated text over different tasks and LLMs discussed in \autoref{ssec:tasks} and \autoref{ssec:models} respectively.

\section{Results \& Discussion}

\subsection{Correlation with Human Evaluation}
Before discussing the benchmark results, we check the effectiveness of the techniques used. We performed human evaluation using the process described in \autoref{ssec:techniques}. The average fact-checking scores using the baseline, 3 techniques, as well domain expert annotations are in \autoref{tab:coorelation-genfact}. The final Cohen’s inter-annotator agreement $\kappa$ is 0.75, which signifies substantial agreement. The baseline technique (FActScore) that uses only Wikipedia as the knowledge source severely underestimates the correctness of the generated text whereas the Chain-of-Thought technique that uses Grounding Document and Wikipedia overestimates it. Overall, it can be seen that our UnVot score derived through joint decisions of CoT and NLI correlates best with domain expert judgment. Still, it is important to point out that this holds true for the summarization, lay summarization, and RAG tasks, while the pure generation task best correlated with baseline FActScore system.

The high correlation of UnVot with human judgment is an important finding. Hiring human annotators, especially domain experts, can be a very expensive and time-consuming process. Having a metric that highly correlates with human scoring intuition can provide a good enough substitute for situations where finding human annotators is infeasible or impossible for certain labs, groups, and application use cases. A lot of focus of recent LLM research is put on aligning LLMs with human values and intuition \citep{wang2023aligning}, and recent LLM-as-judge evaluation metrics like G-Eval \citep{liu-etal-2023-g}, Prometheus \citep{kim2024prometheus}, and TIGERScore \citep{jiang2024tigerscore} put a high emphasis on the correlation of their metrics with humans. As future work, it would be interesting to compare these metrics with UnVot as well, which we currently skip due to resource constraints.

\begin{table}[htpb]
    \centering
      \resizebox{\columnwidth}{!}{%
    \begin{tabular}{lccccc}
        %\toprule
        \textbf{Task} & \textbf{Baseline} &\textbf{CoT*} & \textbf{NLI*} & \textbf{UnVot*} & \textbf{Human}\\
        \toprule
        Summ   & 54.81 & 96.87 & 85.41 & 83.45 & 84.0 \\
        LaySumm  & 52.5 & 97.6 & 91.09 & 88.94 & 88.7 \\
        RAG  & 38.43 & 100.0 &  83.04 & 83.04 & 87.3 \\
        PureGen & 71.26 & 88.17 & 31.61 & 31.31 & 62.7 \\
        \bottomrule
    \end{tabular}
    }
\caption{Fact-checking scores on FActScore (Baseline), Chain-of-Thought (CoT), Natural Language Inference (NLI), Unanimous Voting (UnVot), and Domain Expert Evaluation (Human). * refers to final scores with intrinsic followed by extrinsic fact-checking. }
\label{tab:coorelation-genfact}
\end{table}

\subsection{Task and LLM Performance}

We summarize the Fact-checking scores in \autoref{tab:faithfullness}, which show that the grounding helps LLMs to be more truthful. In terms of tasks, LLMs tend to hallucinate more when prompted to do open-ended generation in the medical domain. However, the performance on other grounding-based task show that given the correct context and supporting document, LLMs are good at understanding a complex domain such as the medical domain. Within each task, LLM performance is mostly uniform. As expected, Open-ended generation is the most challenging task, which is expected due to the LLM using its internal knowledge to answer questions, which can lead to hallucinations. Lay summarization was the most factually correct task, likely owing to the nature of lay text where simpler terms and phrasing is used, which reduces the possibility of mixing up complex scientific terms with one another, which would lead to hallucinations.

\noindent Surprisingly, we see no big difference in models with respect to their sizes. However, both \texttt{Mistral} and \texttt{Mixtral} lead the board for two summarization tasks. While \texttt{Mixtral} performs best for two QA tasks with only the grounding document, \texttt{GPT} comes on top after extrinsic checks, showing its high awareness of Wikipedia in pre-trained knowledge. Two \texttt{Llama} models come close to \texttt{Mixtral}, while \texttt{Gemma} performs the worst on all tasks.

\section{Conclusion and Future Work}
\label{sec:conclusion}
We present a Benchmark providing insights over contemporary LLMs across 4 tasks in the medical domain. We discuss Chain-of-Thought, Natural Language Inference, and Unanimous Voting as fact-checking techniques. Through Domain Expert Evaluation, we show the Unanimous Voting technique to be most reliable. We also explored the effectiveness of two knowledge sources, namely a Grounding Document and Wikipedia, for evaluation and found that using more than one knowledge source leads to an increase in factuality scores. Lastly, we found that LLMs are mostly factually incorrect for Open-ended generation in the medical domain and tend to be more faithful for tasks like Summarization and RAG, where some context is provided to the LLM for generation. We envision our evaluation benchmark to be easily applied for fact-checking across other domains in future.

\section*{Limitation}
Due to the high computation costs, we use only one model as the backbone for each factuality evaluation technique. Even though we evaluated six Large Language Models on four diverse tasks, these tasks may not be enough to capture the entirety of LLM performance and the quickly evolving landscape of new models. 

\noindent Additionally, our two evaluation techniques with NLI and FactScore+ CoT are not perfect and it is possible there were incorrect predictions of which facts were supported or refuted by evidence. Even though our manual inspection and human evaluation showed a good correlation with automated metrics, there will always be some mishaps and incorrect verdicts.

\noindent Finally, our approach relies on making numerous calls to the external API and to the Wikipedia dump database instance in case of extrinsic fact-checking, which can all slow down the overall pipeline. An alternative would have been running locally hosted open-source models, but this was out of our budget due to computational costs. Future work could explore these solutions and make the process faster. 

\section*{Ethics Statement}
Throughout our experiments, we strictly adhere to the ACL Code of Ethics. The manual evaluation was performed by in-house annotators who received a full salary, and their annotation were stored anonymously, mitigating any privacy concerns. They were informed about the task and usability of data in the research. The goal of the research is to evaluate existing techniques and introduce a new technique that can be used for fact-checking LLM generated text on four tasks in the medical domain. We use the LLMs through inference using open-source dataset and do not include in any information in model weights. The discussions and results in this paper are meant to further promote research in the area of LLM Fact-checking as well as create more awareness about their applications in the medical domain. All scripts will be made available to the research community.

\bibliography{custom}

\newpage
\appendix
\section{Technical Details}
\label{sec:appendix-training-details}
\subsection{LLM Generations} 
The inference procedure was done Together AI Inference\footnote{\url{https://www.together.ai/}}. We used the instruct-tuned or chat versions of the models. As for GPT-4o mini, we used the OpenAI API and the latest snapshot available, \texttt{gpt-4o-mini} from Sep 13th, 2024. The checkpoints used for LLM inferences of the open-source models using Together AI are summarized in \autoref{tab:llm-checkpoints}.

\begin{table}[h]
    \centering
      \resizebox{\columnwidth}{!}{%
    \begin{tabular}{lc}
        %\toprule
        \textbf{Model} & \textbf{checkpoint}\\
        \toprule
        \texttt{Llama 3.1 8b}   & meta-llama/Meta-Llama-3.1-8B-Instruct-Turbo \\
        \texttt{Llama 3.1 70b}  & meta-llama/Meta-Llama-3.1-70B-Instruct-Turbo\\
        \texttt{Mistral 7b}  & mistralai/Mistral-7B-Instruct-v0.3 \\
        \texttt{Mixtral 8x7b}   &  mistralai/Mixtral-8x7B-Instruct-v0.1 \\
        \texttt{Gemma 2 9b}  & google/gemma-2-9b-it \\
        
        \bottomrule
    \end{tabular}
}    
\caption{Together AI checkpoints of all LLMs that were used during Inferences. }

\label{tab:llm-checkpoints}
\end{table} 

\subsection{Benchmark Computations}
We used the OpenAI API\footnote{\url{https://platform.openai.com/}} and the latest snapshot available, \texttt{GPT-4o-mini} from Sep 13th, 2024 for Fact-checking using Chain-of-Thought prompting. We leveraged Nvidia V100-16GB and Nvidia A100-80GB GPUs for performing fact-checking. 

\newpage
\paragraph{}
\newpage
\section{LLM Prompts:}
\label{app:prompts}
The prompts used for LLM inferences on all four tasks are illustrated in \autoref{tab:prompt}.
\begin{table*}[!htbp]
\centering
\small
\begin{tabular}{p{15cm}l}
\textbf{TEXT SUMMARIZATION PROMPT}\\
\toprule 
Summarize the given article by including the following key points: \\
\begin{enumerate}
    \item Objective: What is the main research question or objective of the study?
    \item Background: What is the context or rationale for the study?
    \item Methods: What study design, population, and methodologies were used?
    \item Key Findings: What are the most significant results or discoveries from the study?
    \item Conclusions: What conclusions do the authors draw from their findings?
    \item Clinical Relevance: How might the study's findings impact medical practice or patient care?
\end{enumerate}

 Scientific Article: \{article\}  \\
 
 Summary: \\

\\
\textbf{LAY SUMMARIZATION PROMPT}\\
\toprule 

You will be provided a scientific article. Your task is to write a lay summary that accurately conveys the background, methods, key findings and significance of the research in non-technical language understandable to a general audience. Guidelines for crafting a lay summary: \\
\begin{itemize}
    \item Craft a detailed summary that explains the research findings and their implications, providing thorough explanations where necessary.
    \item Ensure factual accuracy and alignment with the research presented in the abstract, elaborating on key points and methodologies.
    \item Highlight the main findings and their implications for real-world scenarios, delving into specific mechanisms or methodologies used in the study and their broader significance.
    \item Incorporate descriptive language to explain complex concepts.
    \item Maintain a balanced tone that is informative and engaging, avoiding technical jargon or overly formal language.
    \item Ensure the summary provides sufficient depth and context to guide the reader through the research journey and address potential questions or areas of confusion.
\end{itemize}

Scientific Article: \{article\} \\
Summary:
 \\
 \\

\textbf{RETRIEVAL AUGMENTED GENERATION PROMPT}\\
\toprule 

Give a simple answer to the question based on the provided context. \\

QUESTION: \{question\} \\

CONTEXT: \{context\}\\

\\

\textbf{OPEN-ENDED GENERATION PROMPT}\\
\toprule 

Give a simple answer to the question based on your best knowledge. \\

QUESTION: \{question\}  \\

\\ 

\end{tabular}
\caption{The prompt in the Benchmark for LLM generation output for all tasks.}
\label{tab:prompt}
\end{table*}

\end{document}